\documentclass[journal]{IEEEtran}
\usepackage{cite}
\usepackage{booktabs}
\usepackage{times}
\usepackage{graphicx}
\usepackage{amsmath}
\usepackage{amssymb}
\usepackage{bbm}
\usepackage{bm}
\usepackage{multirow}
\usepackage{makecell}
\usepackage[table,dvipsnames]{xcolor}
\usepackage[utf8]{inputenc}
\usepackage{xspace}
\usepackage{hyperref}
\hypersetup{colorlinks=true,linkcolor=blue,citecolor=blue,urlcolor=blue}

\newcommand{\figref}[1]{\hyperref[#1]{Fig.~\ref*{#1}}}
\newcommand{\tabref}[1]{\hyperref[#1]{Tab.~\ref*{#1}}}
\newcommand{\secref}[1]{\hyperref[#1]{Sec.~\ref*{#1}}}

\newcommand{\method}{ProMoDD\xspace}

\definecolor{tmmBestCell}{HTML}{DCEEFF}
\definecolor{tmmBestText}{HTML}{1F3A5F}
\definecolor{tmmStdGray}{HTML}{707780}
\newcommand{\result}[2]{%
  \mbox{#1\,{\normalfont\color{tmmStdGray}\tiny $\pm$\,#2}}%
}
\newcommand{\best}[1]{\cellcolor{tmmBestCell}{\color{tmmBestText}\bfseries #1}}
\newcommand{\second}[1]{\underline{#1}}
\newcommand{\na}{N/A}

\newcommand{\tmmTableSetup}{%
  \scriptsize
  \setlength{\tabcolsep}{2.8pt}%
  \renewcommand{\arraystretch}{1.02}%
}

\newenvironment{tmmnotes}{%
  \par\smallskip
  \begin{minipage}{\linewidth}
  \footnotesize
  \setlength{\parindent}{0pt}%
  \setlength{\parskip}{1pt}%
}{%
  \end{minipage}
}

\begin{document}
\bstctlcite{IEEEexample:BSTcontrol}
\raggedbottom
\title{ProMoDD: Dataset Distillation for 3D Human Motion via Prototype-Conditioned Diffusion Models}

\author{Yijin~Wei,
        Yawen~Zou,
        Wenqi~Cai,
        Guang~Li,
        Hangli~Ge,
        Katsuya~Hotta,
        Chunzhi~Gu,
        and~Chao~Zhang%
\thanks{Y.~Wei, Y.~Zou, and W.~Cai are with the Graduate School of Science and Engineering, University of Toyama, Toyama, Japan (e-mail: \nolinkurl{d26c2003@ems.u-toyama.ac.jp}, \nolinkurl{d24c8004@ems.u-toyama.ac.jp}, and \nolinkurl{m25c1055@ems.u-toyama.ac.jp}).}%
\thanks{G.~Li is with the Education and Research Center for Mathematical and Data Science, Hokkaido University, Sapporo, Japan (e-mail: \nolinkurl{guang@lmd.ist.hokudai.ac.jp}).}%
\thanks{H.~Ge is with the Interfaculty Initiative in Information Studies, The University of Tokyo, Tokyo, Japan (e-mail: \nolinkurl{hanglige@g.ecc.u-tokyo.ac.jp}).}%
\thanks{K.~Hotta is with the Faculty of Science and Engineering, Iwate University, Iwate, Japan (e-mail: \nolinkurl{hotta@iwate-u.ac.jp}).}%
\thanks{C.~Gu is with the Faculty of Engineering, University of Fukui, Fukui, Japan (e-mail: \nolinkurl{czgu@ieee.org}).}%
\thanks{C.~Zhang is with the Faculty of Engineering, University of Toyama, Toyama, Japan (e-mail: \nolinkurl{zhang@eng.u-toyama.ac.jp}).}%
\thanks{Corresponding authors: Y.~Wei and C.~Zhang.}
}

\maketitle

\begin{abstract}
Dataset distillation (DD) aims to retain the training utility of a large dataset in a much smaller set of examples. For skeleton-based action recognition, this requires capturing discriminative motion variation without disrupting the geometry and temporal continuity of articulated bodies. Directly optimizing joint coordinates can violate these constraints, while plausible motion generation alone does not ensure useful training examples. We propose Prototype-Conditioned Motion Dataset Distillation (\method), which conditions a motion diffusion model on class-wise prototypes of temporally resolved recognition features. We further tailor diffusion training to motion data, enabling the generator to model both skeletal configurations and their temporal evolution. We then construct each compact training set through budget-aware hybrid selection: representative real sequences cover densely populated regions of the pooled recognition space, while selected generated candidates complement them in less-populated regions. This couples prototype-guided generation with feature-space selection, balancing the reliability of observed motions with the variation of synthesis under a fixed per-class sequence budget. Across five difficulty-stratified subsets of the single-person classes in NTU RGB+D 60, \method achieves its largest gains under tight distillation budgets, reaching $40.32\%$ top-1 accuracy on Full49 with three sequences per class, 4.57 percentage points above class-wise $k$-means. These gains reflect the complete framework. The same distilled sets train ST-GCN, MS-G3D, and CTR-GCN without regeneration.
\end{abstract}


\section{Introduction}
\begin{figure}[!t]
  \centering
  \includegraphics[width=0.98\linewidth]{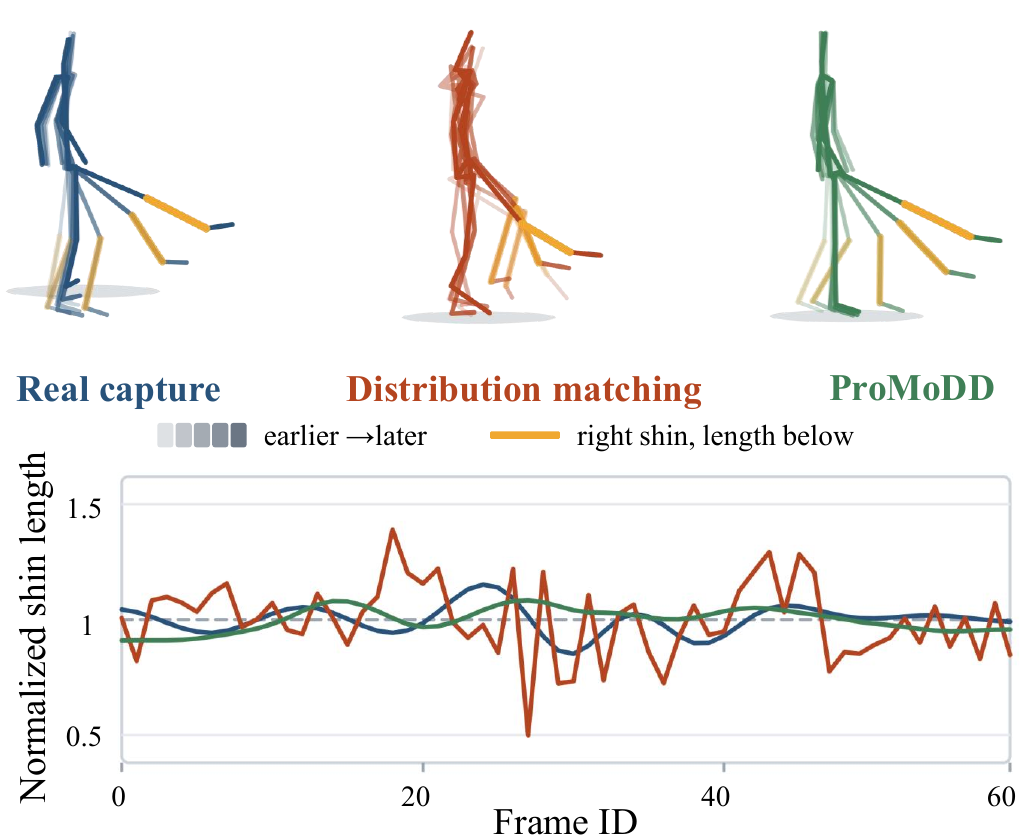}
  \caption{Coordinate-space distribution matching (DM) breaks bone-length consistency, while \method preserves it. In the illustrated sequence, the normalized right-shin length ranges from $0.50$ to $1.39$ across 60 frames under DM, but remains stable in the real and \method sequences.}
  \label{fig:motionbreak}
\end{figure}
Skeleton-based action recognition learns to distinguish activities from sequences of three-dimensional body joints~\cite{sun2022har_review,ren2024skeleton_survey}. Motion collections have grown in scale and diversity~\cite{gupta2021quovadis,cai2022humman,lin2023motionx,zhang2025motionxpp}, from NTU RGB+D~\cite{shahroudy2016ntu} and NTU RGB+D 120~\cite{liu2019ntu120} to broader corpora such as Motion-X~\cite{lin2023motionx}. Although these datasets serve different tasks, their growing scale raises the computational cost of repeatedly training and evaluating motion models. Dataset distillation (DD) seeks to reduce repeated training cost by replacing a large training set with a compact one that remains useful for learning~\cite{wang2018dataset}. For skeleton-based recognition, simply reducing the number of sequences is not enough. The compact set must preserve the motions that distinguish different actions, as well as variations in how the same action is performed. It must also keep the body structure consistent across frames and remain useful for training a classifier.

A major line of DD methods constructs synthetic examples by matching gradients~\cite{zhao2021dc_gradient}, feature distributions~\cite{zhao2023dm_distribution}, or training trajectories~\cite{cazenavette2022mtt} between real and condensed data. Unlike images, skeleton sequences combine two linked structures: joints form an anatomical graph, and their poses evolve over time. A distilled sequence must preserve both valid body geometry and coherent motion while retaining action-discriminative variation. Standard DD objectives do not directly enforce these constraints, making graph-structured temporal data particularly challenging to distill. We see this limitation in a coordinate-space adaptation of distribution matching (DM)~\cite{zhao2023dm_distribution}: in the ``kicking something'' sequence in \figref{fig:motionbreak}, the normalized right-shin length ranges from $0.50$ to $1.39$ across frames, although a skeletal segment should remain approximately constant. This example shows that matching recognition features alone does not guarantee physically coherent motion.

Generative priors help image DD avoid unrestricted pixel optimization~\cite{cazenavette2023glad,su2024d4m}, and motion diffusion offers a learned prior for skeleton sequences~\cite{tevet2023mdm}. Unlike images, however, skeleton motion must preserve both anatomical connections and temporal continuity. We therefore condition diffusion on class-wise prototypes of temporally resolved recognition features, capturing within-class motion patterns beyond action labels alone. A plausible motion is not necessarily a useful training example. Real-only selection is limited to observed motions, whereas an all-synthetic set risks inaccurate or redundant candidates. We thus use budget-aware hybrid construction: representative real sequences cover densely populated regions of the pooled recognition space, while selected generated candidates fill less-populated regions. Because a prototype condition does not determine where a generated sequence lies in the pooled space, we assign candidates using the pooled recognition features of the generated sequences.

Taken together, these requirements call for an integrated framework that connects prototype-guided generation to budget-aware sample selection. We instantiate this framework as Prototype-Conditioned Motion Dataset Distillation (\method). The method centers on a motion diffusion generator tailored to skeleton motion. Conditioned on class-wise prototypes of temporally resolved recognition features, it models skeletal configurations and their temporal evolution. We then combine generated candidates with representative real sequences through budget-aware selection in pooled recognition-feature space. Because generator training is independent of student optimization, it can be reused across sequence budgets and architectures.

To address the lack of an established benchmark for dataset distillation in human motion, we build an evaluation suite from five overlapping, difficulty-stratified subsets of the 49 single-person classes in NTU RGB+D 60. Under the cross-subject protocol~\cite{shahroudy2016ntu}, we evaluate \method across budgets against random selection, class-wise $k$-means, and coordinate-space DM. On Full49 at three sequences per class, \method achieves $40.32\%$ accuracy, 4.57 percentage points above $k$-means. The gap narrows at larger budgets. A same-layer control finds no reliable gain from time-varying conditioning, while the distilled sets train ST-GCN~\cite{yan2018spatial}, MS-G3D~\cite{liu2020disentangling}, and CTR-GCN~\cite{chen2021channel}, without regeneration.

The contributions of this work are summarized as follows.
\begin{itemize}
\item We introduce a motion diffusion model conditioned on class-wise prototypes obtained by clustering temporally resolved recognition features, enabling the generator to capture within-class motion variation.
\item We develop budget-aware hybrid set construction: real samples represent dense pooled-feature regions, while selected generated candidates fill the remaining regions.
\item We establish a difficulty-stratified evaluation suite with five overlapping NTU RGB+D 60 subsets, enabling systematic assessment across distillation budgets and recognition architectures.
\end{itemize}

\section{Related Work}
\noindent\textbf{Motion representations and generation conditions.}
Skeleton-based recognition models learn spatial relations between joints and their temporal evolution. ST-GCN~\cite{yan2018spatial} models the body as a spatiotemporal graph, while MS-G3D~\cite{liu2020disentangling} and CTR-GCN~\cite{chen2021channel} extend the representation with multi-scale dependencies and channel-wise topology refinement. Their intermediate maps retain temporal structure before global aggregation for classification. In motion generation, action-conditioned models use discrete labels to specify the intended activity~\cite{petrovich2021action,guo2020action2motion,zhao2023modiff}. MDM~\cite{tevet2023mdm} and MotionDiffuse~\cite{zhang2022motiondiffuse} extend conditional generation to natural-language descriptions and diffusion-based synthesis~\cite{chen2023mld,zhang2023remodiffuse}. These conditions specify action semantics, but a class label alone cannot distinguish the temporal patterns of individual executions. Our generator instead uses multiple recognition-feature prototypes within each class. The prototypes preserve temporal positions in the extracted representation rather than reducing each motion to one pooled vector. This distinction also separates the objective of our generator from conventional motion synthesis. The aim is not simply to follow a semantic prompt, but to produce sequences that retain information useful for training an action classifier.

\begin{figure*}[!t]
  \centering
  \includegraphics[width=0.85\textwidth]{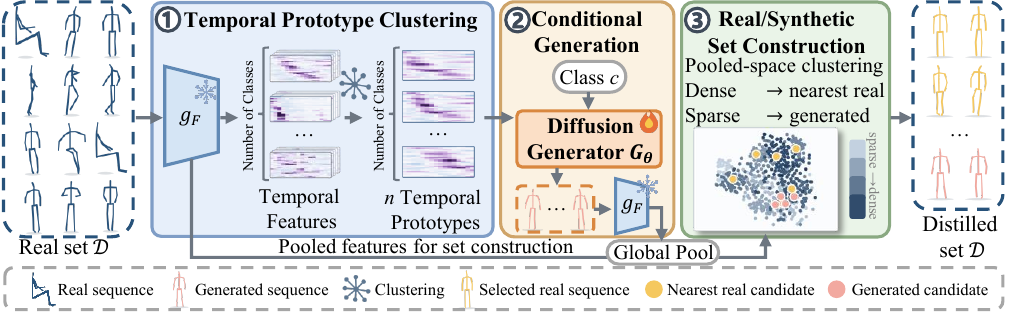}
  \caption{Overview of \method. A frozen recognizer extracts temporally resolved features from real sequences, which are clustered into class-wise prototypes to condition the motion diffusion model. The generated candidates and real sequences are then selected in pooled feature space under a per-class budget. Real sequences represent densely populated regions, while generated candidates are selected for less-populated regions.}
  \label{fig:pipeline}
\end{figure*}

\noindent\textbf{Dataset distillation and generative priors.}
Dataset distillation seeks a small training set that retains the utility of a larger one~\cite{yu2023dd_review}. DC~\cite{zhao2021dc_gradient}, DM~\cite{zhao2023dm_distribution}, and MTT~\cite{cazenavette2022mtt} respectively match training gradients, feature distributions, and optimization trajectories~\cite{wang2022cafe}. When adapted to skeleton coordinates, direct sample optimization can modify joint locations without preserving bone geometry or temporal consistency. Generative approaches introduce learned priors to constrain the space of synthetic data~\cite{zou2025dataset,Cai_2026_CVPR}. GLaD optimizes data through a generative latent representation~\cite{cazenavette2023glad}, while Minimax Diffusion and D$^4$M use diffusion-based generation for compact image training sets~\cite{gu2024minimax,su2024d4m,moser2024ld3m}. UniDD introduces task-guided diffusion for multiple visual tasks, and learnability-guided diffusion adapts generation to the state of the learner~\cite{qi2025universal,chansantiago2026learnability}. ProtoBlend uses feature-space clusters to allocate prototypes in video DD~\cite{ren2026protoblend}. D$^4$M applies diffusion to image latents and uses category prototypes to guide generation~\cite{su2024d4m}. Our method instead uses an ordered sequence of recognition features to guide diffusion over skeleton coordinates. Ignoring temporal order loses information about how an action unfolds, so we preserve it and add losses on motion velocity and acceleration to encourage coherent transitions.

\noindent\textbf{Real-sample selection and hybrid construction.}
Real-data reduction offers a simpler alternative to synthesizing every training entry~\cite{killamsetty2021grad}. Coreset methods select representative recordings, and dataset quantization constructs compact subsets from real images~\cite{sener2018active,zhou2023dataset}. DREAM~\cite{liu2023dream} uses representative samples as targets for more stable dataset matching, whereas RDED~\cite{sun2024diversity} emphasizes realism and diversity in compact data construction. Selection preserves recorded motion but is limited to existing executions. Generation can provide additional candidates, although they may be inaccurate or redundant. We use the two sources together by placing real sequences in densely supported pooled-feature regions and assigning generated candidates to the remaining regions. Unlike DREAM, which uses representatives as matching targets, our retained real sequences are themselves part of the final student training set. Prototype estimation and final-set construction use different feature spaces, allowing temporal conditions to guide generation while pooled features determine which candidates are selected.

\section{Method}
\label{sec:method}

\subsection{Problem Formulation and Overview}
\label{sec:method-overview}
Let $\mathcal{D}=\{(\bm x_i,y_i)\}_{i=1}^{N}$ contain $C$ action classes, with each motion $\bm x_i\in\mathbb R^{F\times J\times3}$. Given a budget of $n$ sequences per class (SPC), we construct a class-balanced surrogate $\widetilde{\mathcal D}=\bigcup_{c=1}^{C}\widetilde{\mathcal D}_c$, where $|\widetilde{\mathcal D}_c|=n$. The goal is to retain recognition utility when a student is trained on this surrogate instead of the full dataset.

\method comprises three stages, as illustrated in \figref{fig:pipeline}. A frozen recognizer provides temporally resolved features, which we cluster within each class to form generation prototypes. These prototypes condition a motion diffusion model to sample skeleton sequences. We select real or generated motion sequences according to their pooled recognition features under the class-wise budget. The features guide selection, while the distilled set contains the corresponding motion sequences. The generator can be reused across budgets and student architectures.

Solving this problem requires class-wise prototypes that capture within-class temporal variation, a motion diffusion model trained to generate coherent skeleton sequences under these conditions, and a budget-aware selector that chooses real or generated motions in a shared recognition-feature space. The following subsections detail these components in order: temporal prototype estimation (\secref{sec:method-prototype}), temporally conditioned skeleton diffusion (\secref{sec:method-generator}), and hybrid set construction (\secref{sec:method-construction}).

\subsection{Recognition-Guided Temporal Prototype Estimation}
\label{sec:method-prototype}

\noindent\textbf{Temporal discriminative representation.}
Let $E_{\mathrm F}$ denote the feature map from an intermediate layer of the frozen CTR-GCN~\cite{chen2021channel} feature extractor $g_{\mathrm F}$. We apply the forward frame-difference transform $\mathcal T$ to each sequence, where $\mathcal T(\bm x_i)_f=\bm x_{i,f+1}-\bm x_{i,f}$ for $f=1,\ldots,F-1$ and $\mathcal T(\bm x_i)_F=\bm 0$. Averaging $E_{\mathrm F}(\mathcal T(\bm x_i))$ over the 22 joints and person dimension yields $\bm Z_i\in\mathbb R^{L\times d_f}$. Subtracting the mean temporal map of the real training set and applying vectorized $L_2$ normalization gives $\bm z_i\in\mathbb R^{L\times d_f}$.

We use an intermediate feature map rather than the final class-level representation of $g_{\mathrm F}$. The final representation is globally aggregated for action classification and therefore lacks frame-wise structure.
The intermediate map retains features at individual temporal positions, which can be encoded as temporal tokens for the diffusion model.
It thus provides recognition-guided temporal cues without directly representing joint trajectories or physical speed. Because we average over joints and persons, it does not indicate which joints contribute to each feature.
This design preserves temporal structure for conditioning, but does not by itself establish an accuracy benefit from temporal resolution.

\begin{figure*}[!t]
  \centering
  \includegraphics[width=0.95\textwidth]{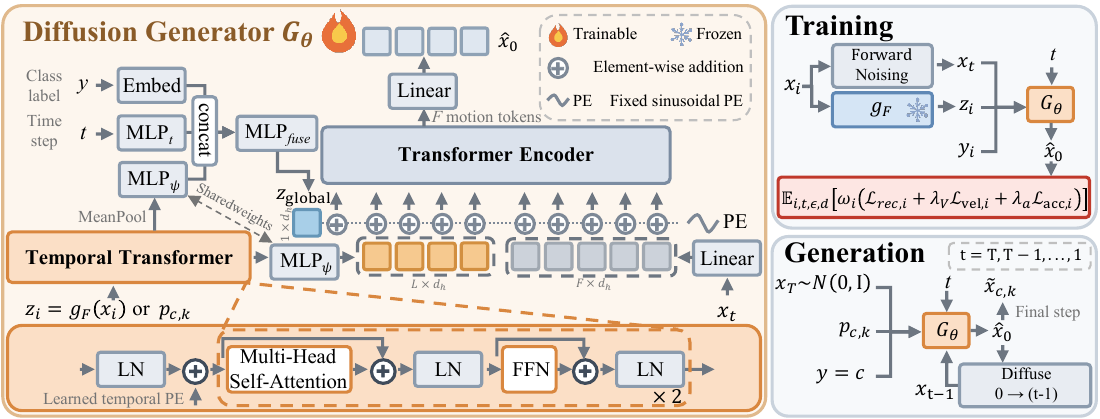}
  \caption{Training and generation of the conditional motion generator. During training, the forward process adds noise to a real motion $\bm x_i$ to obtain $\bm x_t$, while the frozen recognizer extracts a temporal condition $\bm z_i$. Given $(\bm x_t,t,y_i,\bm z_i)$, the generator predicts the clean motion. During generation, the model starts from Gaussian noise, and the class-wise prototype $\bm p_{c,k}$ replaces $\bm z_i$ as the temporal condition. The temporal and global conditions guide denoising.}
  \label{fig:diffusion_generator}
\end{figure*}

\noindent\textbf{Confidence filtering and stability weighting.}
Class-ambiguous sequences can make class prototypes less representative, while bone-length instability can indicate unreliable skeletal geometry. We therefore use teacher confidence to select prototype-estimation samples and stability weights to reduce the influence of unstable motions in prototype estimation and generator training.
For confidence filtering, a separate frozen recognition teacher $f_{\mathrm T}$ assigns each sequence its ground-truth-class probability $\gamma_i=[\operatorname{softmax}(f_{\mathrm T}(\bm x_i))]_{y_i}$. Within each class, we retain the fraction $\kappa$ with the highest confidence for prototype estimation. Generator training continues to use all real training sequences.

For each skeletal edge $b=(p_b,q_b)$, let $\ell_{i,f,b}$ be the Euclidean distance between its endpoint joints in frame $f$. We measure the instability of sequence $i$ by averaging the coefficient of variation over all skeletal edges.
\begin{equation}
 u_i=\frac{1}{|\mathcal B|}\sum_{b\in\mathcal B}
 \frac{\operatorname{Std}_{f}(\ell_{i,f,b})}{\operatorname{Mean}_{f}(\ell_{i,f,b})}.
\label{eq:bone_cv}
\end{equation}
Here $\mathcal B$ is the set of skeletal edges. A lower score indicates more consistent bone lengths across frames. We use this score to define the sample weight
\begin{equation}
\widetilde\omega_i=\exp\!\left(-\frac{\lambda_s u_i}{\operatorname{med}(u)}\right),\qquad
\omega_i=\frac{\widetilde\omega_i}{N^{-1}\sum_{j=1}^{N}\widetilde\omega_j},
\label{eq:bonew}
\end{equation}
Here, $\operatorname{med}(u)$ denotes the median of the sequence-level scores $u_i$ across the complete real training set, computed before confidence filtering. It sets the reference scale for instability. A larger $\lambda_s>0$ penalizes high-$u_i$ motions more strongly, reducing their weight relative to stable motions. We normalize the weights so that their average over the complete real training set is one. In prototype construction, $\omega_i$ weights each retained feature vector in Eq.~\eqref{eq:temporal_prototype}. In generator training, it scales the corresponding per-sequence loss in Eq.~\eqref{eq:gen_objective}. Confidence filtering controls only prototype-clustering membership, and the generator is still trained on all real sequences.

\noindent\textbf{Class-wise temporal prototypes.}
After confidence filtering, we compute class-wise prototypes from the real training features, before training the diffusion generator. Within each class $c$, weighted spherical $k$-means with weighted $k$-means++ initialization~\cite{arthur2007kmeanspp} compares each retained vector $\bm v_i=\operatorname{vec}(\bm z_i)$ with the current cluster centers and assigns it to the most similar center by cosine similarity. Let $\mathcal R_c$ denote the retained indices and $a(i)$ the resulting cluster assignment. Each prototype is the normalized, stability-weighted mean feature vector of a cluster, reshaped to preserve its temporal layout:
\begin{equation}
\bm s_{c,k}=\sum_{\substack{i\in\mathcal R_c\\a(i)=k}}\omega_i\bm v_i,\qquad
\bm p_{c,k}=\operatorname{reshape}_{L\times d_f}\!\left(\frac{\bm s_{c,k}}{\|\bm s_{c,k}\|_2}\right).
\label{eq:temporal_prototype}
\end{equation}
For SPC $=n$, we set $K=n$ and use each prototype as a temporal condition for the diffusion generator. The next subsection describes how these conditions guide motion generation.

\subsection{Temporally Conditioned Skeleton Diffusion}
\label{sec:method-generator}
We build on the $\bm x_0$-prediction formulation of MDM~\cite{tevet2023mdm}. During training, forward noising converts a clean skeleton sequence $\bm x_0=\bm x_i$ into a noisy input $\bm x_t$ at diffusion step $t$. In parallel, frozen $g_{\mathrm F}$ extracts the temporal condition $\bm z_i$. The diffusion generator takes $(\bm x_t,t,y_i,\bm z_i)$ as input and predicts clean motion. During generation, the class prototype $\bm p_{c,k}$ replaces $\bm z_i$, and sampling starts from Gaussian noise (\figref{fig:diffusion_generator}). The generator operates on coordinates without a latent decoder and is trained independently of the downstream student.

\noindent\textbf{Temporal conditioning.}
As illustrated in \figref{fig:diffusion_generator}, a temporal Transformer encodes the feature map used to condition the diffusion generator: it takes the sample feature $\bm z_i$ during training and the class prototype $\bm p_{c,k}$ during generation. Positional encodings preserve the order of the $L$ temporal positions, and the Transformer maps them to $L$ condition tokens. A shared projection maps these tokens and their temporally pooled summary to the generator width $d_h$. The pooled summary is combined with the action-label and diffusion-timestep embeddings to form a global condition token. The generator prepends this global token and the temporal tokens to the projected noisy motion frames, allowing self-attention to use both sequence-level and time-resolved information. The temporal Transformer and diffusion generator are optimized jointly, while the feature extractor $g_{\mathrm F}$ remains frozen.

\noindent\textbf{Motion-aware generator training.}
The diffusion generator predicts clean motion, denoted by $\hat{\bm x}_0$. We use classifier-free guidance (CFG)~\cite{ho2022classifierfree}: during training, the action-label and temporal conditions are jointly dropped with probability $p_{\mathrm{drop}}$, so the same generator learns both conditional and unconditional predictions. At sampling, these predictions are combined using guidance scale $w$. The unconditional prediction serves only as a reference for the conditional prediction. CFG uses their difference to strengthen the influence of the supplied class label and prototype, controlling condition adherence rather than removing conditioning. In addition to coordinate reconstruction, we match the first and second temporal differences of predicted and target motions.
\begin{equation}
\mathcal L_{\mathrm{vel},i}=\frac{\|\Delta\hat{\bm x}_0-\Delta\bm x_0\|_{\mathrm F}^2}{3J(F-1)},\quad
\mathcal L_{\mathrm{acc},i}=\frac{\|\Delta^2\hat{\bm x}_0-\Delta^2\bm x_0\|_{\mathrm F}^2}{3J(F-2)}.
\label{eq:temporal_regularization}
\end{equation}
Here, $\Delta$ and $\Delta^2$ denote first- and second-order differences along the frame axis. For an $F$-frame sequence with $J$ joints and three Cartesian coordinates per joint, these differences contain $3J(F-1)$ and $3J(F-2)$ scalar entries, respectively. Thus, the denominators count the elements in each difference tensor, making both terms mean squared errors. The coordinate reconstruction term $\mathcal L_{\mathrm{rec},i}$ is the mean squared error between $\hat{\bm x}_0$ and $\bm x_0$. The base generator objective is
\begin{equation}
\mathcal L_{\mathrm{gen}}=\mathbb E_{i,t,\bm\epsilon,d}\!\left[\omega_i\big(\mathcal L_{\mathrm{rec},i}+\lambda_v\mathcal L_{\mathrm{vel},i}+\lambda_a\mathcal L_{\mathrm{acc},i}\big)\right].
\label{eq:gen_objective}
\end{equation}
Here, $t$, $\bm\epsilon$, and $d$ denote the diffusion step, Gaussian noise, and condition dropout. The weights $\lambda_v$ and $\lambda_a$ weight the temporal losses. Matching motion derivatives to the real sequence avoids favoring stationary motion. At sampling, Gaussian noise is denoised under $(c,\bm p_{c,k})$ with classifier-free guidance scale $w$.

\subsection{Real and Synthetic Set Construction}
\label{sec:method-construction}
At SPC $=n$, one option is to use the generator to produce all $n$ sequences per class. Generated candidates expand the pool beyond the selected real examples, but may be inaccurate or redundant, which can limit their training value. Selecting only real sequences avoids synthesis errors, but restricts the distilled set to motions observed in the training data. We therefore combine representative real sequences with selected generated candidates, balancing the reliability of real examples with the additional variation offered by synthesis under a fixed per-class budget.

For each class, the generator produces $n$ candidate motions. We set the real-sample fraction to $\rho(n)=\frac{n}{n+n_0}$, allocate $n_{\mathrm{real}}=\operatorname{round}\!\left(\rho(n)n\right)$ sequences to the real subset, and use the remaining $n_{\mathrm{gen}}=n-n_{\mathrm{real}}$ generated candidates. The parameter $n_0>0$ controls how the real-sample fraction increases with the budget, shifting the allocation from generation-dominant at small budgets toward real-sample selection as $n$ grows.

\figref{fig:pipeline} illustrates our density-aware candidate selection strategy. We retain real sequences as reliable representatives of densely populated regions, where observed motions provide strong support, and use selected generated candidates to complement less-populated regions with limited real coverage. This pairs the reliability of observed motions with the potential variation of synthesis. Since prototype conditioning does not determine where generated motions fall in pooled space, we use globally pooled features to compare both sources. For each real sequence, we extract the globally pooled classifier-input feature $\bm r_i\in\mathbb R^{d_f}$ from $g_{\mathrm F}$, subtract the full real-subset mean, and $L_2$-normalize it with a small stabilizer to obtain $\bm h_i$. Generated candidates use the same real-data mean and normalization to obtain $\widetilde{\bm h}_{c,k}$. These pooled features support selection, while temporally resolved features condition generation. Both representations come from the same frozen $g_{\mathrm F}$, avoiding the need to train a separate selector.

For each class, $k$-means partitions real pooled features into $n$ regions, which we rank by occupancy, defined as the number of real training sequences assigned to each region. We select the real sequence nearest to each of the $n_{\mathrm{real}}$ most populated centers, then assign each remaining center in descending occupancy order its nearest unused generated candidate. Formally,
\begin{equation}
\begin{aligned}
 i_{c,j}^{\star}&=\arg\min_{i\in\mathcal I_{c,j}}\|\bm h_i-\bm m_{c,j}\|_2,
 &&j\le n_{\mathrm{real}},\\
 k_{c,j}^{\star}&=\arg\min_{k\notin\mathcal U_c}\|\widetilde{\bm h}_{c,k}-\bm m_{c,j}\|_2,
 &&j>n_{\mathrm{real}},
\end{aligned}
\label{eq:set_assignment}
\end{equation}
Here, $\bm m_{c,j}$ is the pooled-feature center of region $j$ for class $c$, $\mathcal I_{c,j}$ contains its real members, and $\mathcal U_c$ tracks selected generated candidates. Candidates are matched by their pooled features after synthesis and may remain far from a target center if no close match is available.

\section{Experimental Setup}
\label{sec:experimental_setup}

In the absence of an established benchmark for human-motion dataset distillation, we construct an evaluation suite to support controlled comparisons across distillation budgets and action difficulties. We choose NTU RGB+D 60 because it is a widely used, large-scale skeleton-action dataset with broad action coverage and an official cross-subject (CS) protocol, providing a reproducible split and a common reference point~\cite{shahroudy2016ntu}. We focus on its $49$ single-person classes. The CS training and test splits contain $32{,}772$ and $13{,}459$ sequences, respectively.
\subsection{Difficulty-Stratified Evaluation Subsets}
\label{sec:benchmark}
An aggregate Full49 score can hide how well distillation preserves accuracy on easily recognized versus challenging actions. We therefore use difficulty-stratified subsets as descriptive stress tests, while retaining Full49 as the primary evaluation. A reference CTR-GCN trained on Full49 ranks classes by recognition difficulty. The per-example score combines a deficit in ground-truth confidence, a margin-derived term, normalized predictive entropy, and a misclassification indicator, with weights $0.4$, $0.3$, $0.2$, and $0.1$. The margin-derived term is half the sum of the confidence deficit and the largest wrong-class probability. The misclassification indicator is one when the predicted class differs from the ground-truth label and zero otherwise. We compute class difficulty as the weighted sum $S_{\mathrm{class}}(c)=0.7\bar d_c+0.3q_c$, where $\bar d_c$ is the mean per-example score for class $c$ and $q_c=\max_{c'\ne c}M_{c,c'}$ is its largest off-diagonal entry in the row-normalized confusion matrix $M$ (true classes are rows and predicted classes are columns).

Easy10 contains ranks $1$ to $10$, Medium20 ranks $15$ to $34$, Hard20 ranks $30$ to $49$, and Mixed30 ranks $1$ to $10$, $20$ to $29$, and $40$ to $49$. Full49 uses all classes. These subsets overlap and retain all available training sequences for their selected classes. The class ranking is obtained from predictions on the official CS test split. Thus, these subsets are descriptive difficulty-stratified evaluations rather than independent benchmarks. Test predictions determine subset membership only. They are not used for per-sequence selection or as generator-conditioning targets. The number of classes and train/test sequences in each subset is summarized in \tabref{tab:dataset_stats}.

\begin{table}[!t]
\centering
\tmmTableSetup
\caption{Composition of our NTU RGB+D 60 evaluation subsets.}
\label{tab:dataset_stats}
\begin{tabular*}{\columnwidth}{@{\extracolsep{\fill}}lrrrrr@{}}
\toprule
& Easy10 & Medium20 & Hard20 & Mixed30 & Full49 \\
\midrule
\# Classes & 10 & 20 & 20 & 30 & 49 \\
\# Training sequences & 6,682 & 13,394 & 13,372 & 20,045 & 32,772 \\
\# Test sequences & 2,751 & 5,497 & 5,490 & 8,237 & 13,459 \\
\bottomrule
\end{tabular*}
\end{table}

\subsection{Skeleton Preprocessing}
\label{sec:data_preprocessing}
Our generator is defined on a fixed 22-joint HumanML3D skeleton, so we map NTU's 25-joint annotations to this layout to provide a consistent input topology and joint ordering. We retain the first body track and perform the mapping through direct joint correspondences and interpolated spine and collar positions, without SMPL fitting. Sequences with fewer than eight valid frames are discarded. We interpolate isolated joint spikes that move more than $0.5$ m from both adjacent frames, then apply an orthonormal DCT retaining its lowest $K_{\mathrm{DCT}}=10$ coefficients. Motions are pelvis-centered, aligned using the torso and hips, and scaled by mean pelvis-to-neck length. We resample valid frames to $F=60$ for generation.

\subsection{Evaluation Protocol and Metrics}
\label{sec:evaluation_protocol}\label{sec:metrics}
\noindent\textbf{Comparison protocol.} We evaluate SPC values of $1,3,5,10,20,30,50$, and $100$. Each method uses the same number of training entries per class. Baselines are random real selection, class-wise pooled-feature $k$-means, and coordinate-space DM~\cite{zhao2023dm_distribution}. Random selection uniformly samples $n$ distinct real training sequences per class without replacement. For the class-wise $k$-means baseline, we fit $n$ clusters to the normalized pooled features $\bm h_i$ defined in \secref{sec:method-construction} and select the real sequence nearest to each cluster center. Our DM implementation initializes $n$ synthetic sequences per class from randomly sampled real examples. Their joint coordinates are optimized to match class-wise real-data feature means using a CTR-GCN randomly reinitialized each iteration, without explicit skeletal-geometry or temporal-consistency constraints. Student architecture, optimizer, augmentations, epochs, and the training-time matching (TTM) objective~\cite{su2024d4m} are matched within comparisons. We evaluate each method using five seeds $s\in\{0,\ldots,4\}$ and report the mean and standard deviation of test-set top-1 accuracy, our primary comparison metric. For \method, each seed jointly controls diffusion sampling, pooled-space $k$-means initialization, and student training, including initialization, data order, and augmentation. The corresponding random operations are also seeded for the selection and DM baselines.

\noindent\textbf{Training and seed control.} The feature extractor is trained once on Full49. Each subset-specific teacher and each generator configuration are trained once and held fixed across the five evaluation seeds. All generator training runs use seed 10. Thus, the reported standard deviations reflect variability in compact-set construction and student training, but not variability from independently retrained generators. All compact sets receive the same teacher-assisted student training.

\noindent\textbf{Motion metrics.} We also report bone-length variation (BLV) as $100$ times the mean instability score in Eq.~\eqref{eq:bone_cv}. Jerk is calculated from third-order frame differences and averaged over all joints, frames, and sequences. Normalized jerk (NJ) divides this value by the jerk of the complete real Full49 training split. Both diagnostics use the stored $60$-frame, $22$-joint sequences before augmentation, without further smoothing. The differences are computed over the resampled frame index, so NJ is a discrete sequence statistic rather than physical jerk in SI units. BLV summarizes all $21$ bones, unlike the single-shin plot in \figref{fig:motionbreak}.

\subsection{Implementation Details}
\label{sec:implementation}
\noindent\textbf{Models and features.} \tabref{tab:training_config} summarizes the settings. The Full49 feature extractor $g_{\mathrm F}$ is trained once and reused across subsets for temporal conditioning and pooled-feature selection. The subset-specific teacher $f_{\mathrm T}$ provides confidence scores for prototype filtering and soft targets for student training. Both use CTR-GCN~\cite{chen2021channel}. No external pretraining is used, and ST-GCN and MS-G3D~\cite{yan2018spatial,liu2020disentangling} are also evaluated on the same distilled sets. For feature extraction, $g_{\mathrm F}$ runs in evaluation mode, with $E_{\mathrm F}$ extracted after residual addition and ReLU in the eighth residual block, whose module name is \texttt{l8}. Strides in blocks 5 and 8 reduce the temporal dimension from 64 to 16. \textbf{Student training.} Following D4M~\cite{su2024d4m}, all methods use the same TTM objective and student settings. TTM combines label supervision with KL matching to the frozen teacher, which receives the same augmented or mixed inputs as the student. Generated samples use their conditioning-class labels. Augmentation uses rigid rotations and temporal and joint-dimension mixing~\cite{xiang2024joint}. The KL term uses temperature $\tau$ and is scaled by $\tau^2$, while label supervision uses unit temperature. The coefficients $\alpha$ and $1-\alpha$ weight the respective terms. \textbf{Generator optimization.} The objective in Eq.~\eqref{eq:gen_objective} uses stability weights normalized by their minibatch mean. Gradients are clipped to an $\ell_2$ norm of $1.0$, and updates with non-finite gradients are skipped. \textbf{Hardware.} Experiments were run on a system with an Intel Core Ultra 9 CPU, 64.0 GiB of memory, and two RTX 5090 GPUs. 

\begin{table}[!t]
\centering
\caption{Default experimental settings.}
\label{tab:training_config}
\footnotesize
\setlength{\tabcolsep}{1.4pt}
\renewcommand{\arraystretch}{1.0}
\begin{tabular*}{\columnwidth}{@{\extracolsep{\fill}}lr@{\hspace{4pt}}lr@{}}
\toprule
\multicolumn{2}{c}{Student} & \multicolumn{2}{c}{Generator} \\
\cmidrule(r){1-2}\cmidrule(l){3-4}
Optimizer & Nesterov SGD & Optimizer & AdamW \\
LR & $0.05$ & LR & $10^{-4}$ \\
Batch & 16 & Batch & 256 \\
Training & 80 epochs & Training & 160k steps \\
Momentum & $0.9$ & Layers / width & 8 / 512 \\
Weight decay & $4\times10^{-4}$ & CFG dropout & $0.1$ \\
Warm-up & 5 epochs & Diffusion & 1000, cosine \\
LR decay & $\times0.1$ at 50, 70 & Sampling & 1000 steps \\
$\tau$, $\alpha$ & $4$, $0.5$ & CFG scale $w$ & $2.5$ \\
\midrule
\multicolumn{4}{c}{Recognition and prototypes} \\
\midrule
MS-G3D scales & 12 & Retention $\kappa$ & $0.8$ \\
Feature block & 8 & Mixing $n_0$ & 7 \\
Temporal tokens $L$ & 16 & Feature dim. $d_f$ & 256 \\
$\lambda_s$ & 1 & $\lambda_v$, $\lambda_a$ & $1$, $1$ \\
\bottomrule
\end{tabular*}
\end{table}

\section{Experiments}
\label{sec:experiments}

\subsection{Comparative Results}
\label{sec:comparative}

\begin{table*}[!t]
\centering
\small
\setlength{\tabcolsep}{2pt}
\renewcommand{\arraystretch}{1.0}
\caption{Main results on the NTU RGB+D 60 single-person subsets under the
cross-subject (CS) protocol. Entries report top-1 accuracy (\%) as mean $\pm$
standard deviation over five evaluation seeds. The trained generator is
held fixed.}
\label{tab:main_cs}
\resizebox{0.7\textwidth}{!}{%
\begin{tabular}{llrrrrrrrr}
\toprule
\makecell{Subset\\{\normalfont\tiny (Full data)}} & Method
& \multicolumn{1}{c}{SPC${=}1$}
& \multicolumn{1}{c}{SPC${=}3$}
& \multicolumn{1}{c}{SPC${=}5$}
& \multicolumn{1}{c}{SPC${=}10$}
& \multicolumn{1}{c}{SPC${=}20$}
& \multicolumn{1}{c}{SPC${=}30$}
& \multicolumn{1}{c}{SPC${=}50$}
& \multicolumn{1}{c}{SPC${=}100$} \\
\midrule
\multirow{4}{*}{\makecell[l]{Easy10\\{\normalfont\color{tmmStdGray}\tiny 98.98\%}}} & Random & \best{\result{11.78}{2.62}} & \result{\second{71.44}}{4.35} & \result{\second{87.55}}{1.42} & \result{92.13}{0.55} & \best{\result{95.33}{0.49}} & \result{\second{96.29}}{0.27} & \result{97.16}{0.06} & \result{98.10}{0.20} \\
 & $k$-means & \best{\result{11.78}{2.62}} & \result{70.79}{1.96} & \best{\result{88.11}{1.19}} & \best{\result{93.06}{0.85}} & \best{\result{95.33}{0.42}} & \result{96.21}{0.28} & \result{\second{97.29}}{0.27} & \best{\result{98.31}{0.23}} \\
 & DM & \best{\result{11.78}{2.62}} & \result{14.09}{3.85} & \result{24.02}{7.52} & \result{45.65}{8.55} & \result{79.60}{4.07} & \result{84.85}{4.03} & \result{85.83}{3.93} & \result{83.25}{3.22} \\
 & \method & \best{\result{11.78}{2.62}} & \best{\result{72.72}{3.99}} & \result{87.46}{0.71} & \result{\second{92.48}}{0.40} & \result{\second{95.06}}{0.31} & \best{\result{96.36}{0.41}} & \best{\result{97.33}{0.20}} & \result{\second{98.18}}{0.18} \\
\midrule
\multirow{4}{*}{\makecell[l]{Medium20\\{\normalfont\color{tmmStdGray}\tiny 87.05\%}}} & Random & \result{9.12}{1.27} & \result{22.89}{2.20} & \result{35.97}{1.78} & \result{55.76}{2.81} & \result{69.27}{0.30} & \result{75.05}{0.67} & \result{79.24}{0.36} & \best{\result{83.48}{0.27}} \\
 & $k$-means & \result{\second{12.88}}{0.96} & \result{\second{29.54}}{2.33} & \result{\second{45.47}}{2.29} & \result{\second{63.57}}{1.07} & \result{\second{72.09}}{1.32} & \best{\result{76.04}{0.70}} & \result{\second{79.56}}{0.67} & \result{82.96}{0.39} \\
 & DM & \result{5.18}{0.33} & \result{6.38}{0.88} & \result{7.56}{1.37} & \result{14.67}{1.24} & \result{22.47}{3.39} & \result{25.56}{3.68} & \result{27.27}{3.09} & \result{31.10}{4.27} \\
 & \method & \best{\result{17.35}{1.75}} & \best{\result{35.71}{1.89}} & \best{\result{50.32}{1.31}} & \best{\result{64.02}{0.99}} & \best{\result{72.26}{0.58}} & \result{\second{75.73}}{0.74} & \best{\result{80.21}{0.29}} & \result{\second{83.00}}{0.31} \\
\midrule
\multirow{4}{*}{\makecell[l]{Hard20\\{\normalfont\color{tmmStdGray}\tiny 68.00\%}}} & Random & \result{7.97}{0.76} & \result{17.52}{1.09} & \result{25.14}{1.75} & \result{33.76}{1.12} & \result{47.40}{0.28} & \result{51.70}{0.42} & \result{56.79}{0.63} & \result{\second{63.18}}{0.40} \\
 & $k$-means & \result{\second{12.45}}{1.07} & \result{\second{22.84}}{1.20} & \result{\second{31.74}}{0.91} & \result{\second{41.90}}{1.16} & \result{\second{49.82}}{0.69} & \result{\second{54.29}}{0.80} & \result{\second{59.39}}{0.85} & \result{62.69}{0.69} \\
 & DM & \result{5.02}{0.03} & \result{5.94}{0.43} & \result{6.59}{0.40} & \result{9.12}{1.14} & \result{11.67}{1.24} & \result{18.81}{3.24} & \result{18.68}{4.18} & \result{23.75}{5.64} \\
 & \method & \best{\result{14.44}{1.11}} & \best{\result{27.61}{3.29}} & \best{\result{35.03}{1.53}} & \best{\result{42.96}{0.51}} & \best{\result{51.14}{0.80}} & \best{\result{55.12}{0.92}} & \best{\result{59.41}{0.46}} & \best{\result{63.91}{0.81}} \\
\midrule
\multirow{4}{*}{\makecell[l]{Mixed30\\{\normalfont\color{tmmStdGray}\tiny 78.43\%}}} & Random & \result{17.02}{1.69} & \result{41.65}{2.20} & \result{49.24}{1.57} & \result{59.29}{1.43} & \result{67.11}{0.84} & \result{70.57}{0.58} & \result{74.12}{0.33} & \best{\result{77.68}{0.44}} \\
 & $k$-means & \result{\second{21.91}}{1.17} & \result{\second{46.23}}{1.45} & \result{\second{53.88}}{0.99} & \result{\second{62.18}}{0.70} & \result{\second{68.06}}{1.08} & \best{\result{71.71}{0.93}} & \result{\second{74.59}}{0.36} & \result{77.00}{0.90} \\
 & DM & \result{3.91}{0.86} & \result{5.09}{0.58} & \result{12.89}{1.73} & \result{23.30}{4.63} & \result{32.72}{3.73} & \result{35.14}{1.25} & \result{34.73}{3.60} & \result{40.95}{2.26} \\
 & \method & \best{\result{23.35}{1.52}} & \best{\result{48.55}{1.22}} & \best{\result{54.23}{0.70}} & \best{\result{62.68}{1.23}} & \best{\result{68.43}{0.67}} & \result{\second{71.69}}{0.30} & \best{\result{75.17}{0.63}} & \result{\second{77.53}}{0.49} \\
\midrule
\multirow{4}{*}{\makecell[l]{Full49\\{\normalfont\color{tmmStdGray}\tiny 76.06\%}}} & Random & \result{14.84}{1.05} & \result{30.38}{1.12} & \result{40.05}{2.41} & \result{51.56}{2.12} & \result{61.98}{0.96} & \result{66.70}{1.01} & \result{70.87}{0.37} & \best{\result{74.70}{0.75}} \\
 & $k$-means & \result{\second{17.18}}{0.74} & \result{\second{35.75}}{0.73} & \result{\second{45.21}}{2.50} & \result{\second{56.31}}{0.72} & \result{\second{63.92}}{1.15} & \result{\second{67.78}}{0.42} & \result{\second{71.62}}{0.32} & \result{74.08}{0.92} \\
 & DM & \result{2.07}{0.03} & \result{3.11}{0.32} & \result{5.38}{2.07} & \result{12.49}{1.57} & \result{20.69}{1.48} & \result{22.32}{2.19} & \result{22.33}{2.26} & \result{24.78}{2.78} \\
 & \method & \best{\result{20.65}{0.91}} & \best{\result{40.32}{0.82}} & \best{\result{47.90}{2.57}} & \best{\result{56.43}{1.75}} & \best{\result{64.57}{1.17}} & \best{\result{68.59}{0.56}} & \best{\result{72.11}{0.46}} & \result{\second{74.30}}{0.91} \\
\bottomrule
\end{tabular}%
}
\end{table*}

\noindent\textbf{Low-budget performance.} \tabref{tab:main_cs} compares compact sets at matched SPC and under the same student-training protocol. At SPC $=3$, \method improves over class-wise $k$-means on all five subsets, with gains of $1.93$ to $6.17$ percentage points. On Full49, accuracy rises from $35.75\%$ to $40.32\%$. The Full49 gap narrows to $0.12$ points at SPC $=10$ and $0.49$ points at SPC $=50$. Thus, \method's clearest advantage appears when only a few sequences can be retained per class.

\noindent\textbf{High-budget performance.} Real-data selection becomes competitive as the budget grows. At SPC $=100$, random selection exceeds $k$-means on four subsets. On Full49, random selection, $k$-means, and \method reach $74.70\%$, $74.08\%$, and $74.30\%$, respectively. With more entries, real-sample methods can retain more recorded executions and approach the full-data reference.

\noindent\textbf{Effect of action difficulty.} Difficulty affects how much accuracy each subset retains under distillation and how much it benefits from synthesis. At SPC $=3$, \method exceeds class-wise $k$-means by $1.93$ points on Easy10, $6.17$ on Medium20, and $4.77$ on Hard20. The gain is therefore larger on Medium20 and Hard20 than on Easy10, although it does not increase monotonically with difficulty. Hard20 also loses more accuracy under distillation: at SPC $=10$, it retains about $63\%$ of its full-data reference accuracy, compared with about $93\%$ for Easy10. One possible explanation is that a few real examples can represent easily recognized actions more effectively, while generated candidates can complement underrepresented regions for some more challenging actions. This interpretation is consistent with our hybrid selection strategy, but the experiments do not establish difficulty as the cause of the gains. The subsets overlap and were ranked using full-data teacher predictions, so these comparisons are descriptive rather than controlled tests of difficulty alone.

\noindent\textbf{Direct coordinate optimization.} Our coordinate-space adaptation of DM performs poorly in the tested setup: on Full49, the frozen teacher recognizes fewer than $6\%$ of the optimized sequences, compared with over $95\%$ of the real initializations, and \figref{fig:motionbreak} shows distorted bone lengths. This result does not show that DM is unsuitable for skeleton-based dataset distillation. Rather, it highlights the difficulty of directly optimizing a sequence of joint coordinates. Unlike image pixels, joint positions are coupled by skeletal geometry and temporal continuity, so unconstrained updates can produce invalid or unrecognizable motions. 

\noindent\textbf{Computational Efficiency.}
\label{sec:efficiency}

\begin{table}[!t]
\centering
\tmmTableSetup
\caption{Construction cost, repeated student cost, storage, and accuracy on Full49 at SPC $=10$ (one RTX 5090).}
\label{tab:efficiency}
\resizebox{0.7\columnwidth}{!}{%
\begin{tabular}{lrrrr}
\toprule
\multirow{2}{*}{Method} & Offline & Student & Storage & Top-1 \\
 & (GPU-h) & (GPU-h) & (GB) & (\%) \\
\midrule
Full real data & \na & 2.89 & 0.5320 & 76.06 \\
Random & $<0.01$ & 0.10 & 0.0080 & \result{51.56}{2.12} \\
$k$-means & $<0.01$ & 0.10 & 0.0080 & \result{\second{56.31}}{0.72} \\
DM & 0.98 & 0.10 & 0.0080 & \result{12.49}{1.57} \\
\method & 4.56 & 0.10 & 0.0080 & \best{\result{56.43}{1.75}} \\
\bottomrule
\end{tabular}%
}
\begin{tmmnotes}
Offline excludes Full49 feature-extractor training ($0.47$ GPU-h) and teacher training. Storage covers coordinates and labels only. 
\end{tmmnotes}
\end{table}

\begin{table*}[!t]
\centering
\tmmTableSetup
\caption{Cross-architecture generalization of \method on the NTU RGB+D 60
single-person evaluation subsets (top-1 accuracy, \%, mean $\pm$ standard
deviation over five seeds). Seeds follow the protocol in
Section~\ref{sec:evaluation_protocol}.}
\label{tab:crossarch}
\resizebox{\textwidth}{!}{%
\begin{tabular}{l*{15}{c}}
\toprule
\multirow{2}{*}{Architecture} & \multicolumn{3}{c}{Easy10}
& \multicolumn{3}{c}{Medium20} & \multicolumn{3}{c}{Hard20}
& \multicolumn{3}{c}{Mixed30} & \multicolumn{3}{c}{Full49} \\
\cmidrule(lr){2-4}\cmidrule(lr){5-7}\cmidrule(lr){8-10}
\cmidrule(lr){11-13}\cmidrule(lr){14-16}
& 3 & 10 & 30 & 3 & 10 & 30 & 3 & 10 & 30
& 3 & 10 & 30 & 3 & 10 & 30 \\
\midrule
ST-GCN (3.08M)
& \result{70.59}{3.61} & \result{92.38}{1.35} & \best{\result{96.50}{0.38}}
& \result{\second{37.51}}{2.76} & \result{59.72}{2.17} & \result{74.84}{0.19}
& \result{21.23}{2.18} & \result{38.49}{1.15} & \result{52.82}{0.85}
& \result{46.30}{2.21} & \result{61.92}{0.81} & \result{70.83}{0.71}
& \result{39.69}{1.64} & \result{\second{56.60}}{1.01} & \result{67.72}{0.53} \\
MS-G3D (3.04M)
& \result{\second{72.08}}{4.59} & \best{\result{92.91}{0.53}} & \result{95.67}{2.25}
& \best{\result{38.97}{2.03}} & \result{\second{62.48}}{0.82} & \result{\second{74.87}}{0.59}
& \result{\second{24.75}}{0.96} & \result{\second{42.19}}{1.12} & \result{\second{54.44}}{0.51}
& \result{\second{48.37}}{0.77} & \best{\result{62.83}{0.39}} & \best{\result{72.40}{0.74}}
& \best{\result{40.41}{0.78}} & \best{\result{57.81}{1.13}} & \best{\result{69.07}{0.44}} \\
CTR-GCN (1.43M)
& \best{\result{72.72}{3.99}} & \result{\second{92.48}}{0.40} & \result{\second{96.36}}{0.41}
& \result{35.71}{1.89} & \best{\result{64.02}{0.99}} & \best{\result{75.73}{0.74}}
& \best{\result{27.61}{3.29}} & \best{\result{42.96}{0.51}} & \best{\result{55.12}{0.92}}
& \best{\result{48.55}{1.22}} & \result{\second{62.68}}{1.23} & \result{\second{71.69}}{0.30}
& \result{\second{40.32}}{0.82} & \result{56.43}{1.75} & \result{\second{68.59}}{0.56} \\
\bottomrule
\end{tabular}%
}
\end{table*}

The offline and repeated-training costs are reported separately in \tabref{tab:efficiency}. On Full49 at SPC $=10$, constructing a \method set costs $4.56$ GPU-h. Training a student on the compact set then takes $0.10$ GPU-h per run, compared with $2.89$ GPU-h on full data. Storage decreases from $0.532$ GB to $0.008$ GB. With the reported costs, the construction expense is offset after about two student runs. This estimate excludes the shared feature-extractor and teacher training. Although class-wise $k$-means is cheaper to construct, \method yields more accurate distilled sets when the per-class budget is tight. On Full49, it improves over $k$-means from $17.18\%$ to $20.65\%$ at SPC $=1$ and from $35.75\%$ to $40.32\%$ at SPC $=3$. At SPC $=3$, this advantage appears on all five subsets, with gains of $1.93$ to $6.17$ points. At SPC $=10$, the methods reach similar accuracy ($56.43\%$ for \method and $56.31\%$ for $k$-means), showing that the real-data baseline becomes competitive at this budget. After construction, both methods have the same student-training cost and storage at equal SPC, and the distilled sets can be reused across student runs. Storage covers coordinates and labels, excluding generator and teacher parameters.

\label{sec:crossarch}
\noindent\textbf{Cross-Architecture Transfer.} As shown in \tabref{tab:crossarch}, the same CTR-GCN-guided distilled sets train ST-GCN, MS-G3D, and CTR-GCN without regenerating samples. On Full49 at SPC $=3$, the three architectures achieve $39.69\%$, $40.41\%$, and $40.32\%$ accuracy, respectively. Their similar performance indicates that the utility of the distilled data is not tied to a specific network architecture. Accuracy varies more on the challenging Hard20 subset ($21.23\%$, $24.75\%$, and $27.61\%$), showing that backbone choice can still affect performance. 

\subsection{Ablation Studies}
\label{sec:ablation}

We examine what the temporal condition contributes, how motion diagnostics relate to training utility, and how source allocation affects recognition.

\begin{table}[!t]
\centering
\tmmTableSetup
\caption{Component ablation on Full49. }
\label{tab:component_ablation}
\resizebox{0.65\columnwidth}{!}{%
\begin{tabular}{lccrrr}
\toprule
Cond. & TT & CF & SPC${=}3$ & SPC${=}10$ & SPC${=}30$ \\
\midrule
$\times$ & $\times$ & $\times$ & \result{33.42}{1.87} & \result{\second{56.26}}{1.20} & \result{67.81}{0.52} \\
fc & $\times$ & $\times$ & \result{34.08}{1.13} & \result{54.54}{1.37} & \result{67.82}{0.49} \\
l8 & $\times$ & $\times$ & \result{37.05}{0.93} & \result{55.84}{0.92} & \result{68.10}{0.63} \\
l8 & $\checkmark$ & $\times$ & \result{\second{40.22}}{1.97} & \result{55.83}{0.95} & \result{68.16}{0.84} \\
l8 & $\times$ & $\checkmark$ & \result{37.22}{2.39} & \result{54.99}{1.56} & \result{\second{68.44}}{0.29} \\
l8 & $\checkmark$ & $\checkmark$ & \best{\result{40.32}{0.82}} & \best{\result{56.43}{1.75}} & \best{\result{68.59}{0.56}} \\
\bottomrule
\end{tabular}%
}
\begin{tmmnotes}
All variants use action labels. fc: pooled classifier-input
features, l8: temporal features extracted after residual addition and ReLU in the eighth residual block, TT: Temporal Transformer, CF: confidence filtering.
\end{tmmnotes}
\end{table}

\begin{table}[!t]
\centering
\tmmTableSetup
\caption{Same-layer conditioning control on Full49 at SPC $=3$. Each variant uses a separately trained generator.}
\label{tab:samelayer_control}
\begin{tabular}{lc}
\toprule
Layer-8 condition & Top-1 (\%) \\
\midrule
Time-resolved & \best{\result{37.67}{1.00}} \\
Time-pooled (fixed assignments) & \result{36.23}{1.90} \\
\bottomrule
\end{tabular}
\begin{tmmnotes}
Both conditions use layer-8 features and the same prototype assignments. Time-resolved retains the 16 ordered feature tokens. Time-pooled averages them over time, L2-normalizes the mean, and repeats it at all 16 positions.
\end{tmmnotes}
\end{table}

\noindent\textbf{Conditioning component ablation.} On Full49 at SPC $=3$, replacing pooled classifier-input features with temporally resolved features from the eighth residual block (l8) raises accuracy from $34.08\%$ to $37.05\%$ (\tabref{tab:component_ablation}). With the block-$8$ features, adding the Temporal Transformer raises accuracy to $40.22\%$, and confidence filtering further increases it to $40.32\%$. At SPC $=30$, the accuracy range across conditioning variants is less than one point. These results show that the complete conditioning pipeline is most beneficial at a tight distillation budget. The separate same-layer control below examines the contribution of temporal resolution more directly.

\noindent\textbf{Time-resolved condition ablation.} To test whether retaining temporal variation itself benefits distilled-set utility, rather than merely comparing different feature depths, we compare time-resolved and time-pooled conditions using the same layer-8 features and prototype assignments (\tabref{tab:samelayer_control}). At Full49 SPC $=3$, time-resolved conditioning achieves higher mean top-1 accuracy than time-pooled conditioning ($37.67\%$ vs. $36.23\%$) and lower standard deviation across five evaluation seeds ($1.00$ vs. $1.90$, respectively). Thus, it performs better and more consistently in this comparison.

\begin{table}[!t]
\centering
\tmmTableSetup
\caption{Motion-regularization ablation on Full49.}
\label{tab:regularization}
\resizebox{0.9\columnwidth}{!}{%
\begin{tabular}{lccccc}
\toprule
& \multicolumn{2}{c}{Motion diagnostics}
& \multicolumn{3}{c}{Student top-1 (\%)} \\
\cmidrule(lr){2-3}\cmidrule(lr){4-6}
\raisebox{1.7ex}[0pt][0pt]{Configuration} & BLV\,(\%) & NJ (real${=}1$)
& SPC${=}3$ & SPC${=}10$ & SPC${=}30$ \\
\midrule
No regularization
& 7.55 & 2.925 & \result{37.58}{2.20} & \result{55.29}{0.99} & \result{67.89}{0.48} \\
Temporal losses
& 6.74 & 2.059 & \result{37.54}{1.03} & \result{55.17}{1.34} & \second{\result{68.22}{0.22}} \\
Stability weighting
& 6.70 & 2.446 & \result{38.15}{1.40} & \result{54.78}{1.32} & \result{67.76}{0.64} \\
Full model
& \second{5.50} & \best{1.672} & \best{\result{40.32}{0.82}}
& \best{\result{56.43}{1.75}} & \best{\result{68.59}{0.56}} \\
Full model $+\mathcal{L}_{\mathrm{bone}}$
& \best{5.36} & \second{1.761} & \second{\result{39.27}{1.22}} & \second{\result{55.77}{1.19}} & \result{68.18}{0.41} \\
\midrule
Real data & 7.04 & 1.000 & \multicolumn{3}{c}{\na} \\
\bottomrule
\end{tabular}%
}
\begin{tmmnotes}
BLV denotes temporal bone-length variation. NJ denotes jerk normalized by the
real-data value. Motion diagnostics use 1,470 generated sequences per
configuration (30 per class, generation seed 0). Student accuracy follows the
five-seed protocol in Section~\ref{sec:evaluation_protocol}. Best and
second-best values are marked column-wise among generated configurations.
\end{tmmnotes}
\end{table}

\noindent\textbf{Motion-regularization ablation.} To test whether regularizers that improve motion diagnostics also improve distilled-set utility, we compare temporal losses, stability weighting, their combination, and an additional direct bone-length loss (\tabref{tab:regularization}). Temporal losses reduce normalized jerk (NJ) from $2.925$ to $2.059$, with little change in SPC $=3$ accuracy ($37.58\%$ to $37.54\%$), while stability weighting mainly reduces bone-length variation (BLV) from $7.55\%$ to $6.70\%$. Combining both yields $40.32\%$ accuracy, BLV of $5.50\%$, and NJ of $1.672$. Adding the direct bone-length loss lowers BLV further to $5.36\%$, but accuracy drops to $39.27\%$, showing that better motion diagnostics do not necessarily mean greater training utility.

\noindent\textbf{Real--synthetic assignment ablation.} To test whether the location of real and generated samples matters beyond their ratio, we compare dense-first and reversed sparse-first assignment (\tabref{tab:density}). For each class, we cluster the normalized pooled features of real training sequences into $n$ regions ($n$ is the per-class budget) and rank them by occupancy, or the number of real sequences in each region. The $n_{\mathrm{real}}$ highest-occupancy regions are dense, and the remaining $n_{\mathrm{gen}}$ are sparse. Given the same pool of $n$ generated candidates (one per temporal prototype), dense-first selects the nearest real sequence for each dense-region center and the nearest unused generated candidate for each sparse-region center. The reverse policy selects real sequences nearest to the $n_{\mathrm{real}}$ lowest-occupancy centers and assigns generated candidates to the rest with the same nearest-unused rule. We hold the partition, source counts, candidate pool, and student seeds fixed. Dense-first wins in 17 of 20 settings across Medium20, Hard20, Mixed30, and Full49, by $1.17$ points on average; Easy10 slightly favors the reverse. Regions assigned to real samples contain $73\%$ and $93\%$ of the real training sequences at SPC $=10$ and $30$, respectively. Varying the tested real-to-generated ratio changes Full49 SPC $=10$ accuracy by at most $0.64$ points. These results are consistent with real examples anchoring well-supported regions and generated samples complementing less-populated ones, although occupancy does not directly measure generated-sample coverage or motion novelty.

\begin{table}[!t]
\centering
\tmmTableSetup
\caption{Real–synthetic assignment ablation.}
\label{tab:density}
\resizebox{\columnwidth}{!}{%
\begin{tabular}{llccccc}
\toprule
Subset & \multicolumn{1}{c}{Setting} & SPC${=}3$ & SPC${=}5$ & SPC${=}10$ & SPC${=}20$ & SPC${=}30$ \\
\midrule
\multirow{2}{*}{Easy10} & dense-first (ours) & \result{72.72}{3.99} & \result{87.46}{0.71} & \result{92.48}{0.40} & \result{95.06}{0.31} & \result{96.36}{0.41} \\
 & sparse-first (reverse) & \best{\result{73.72}{2.96}} & \best{\result{87.70}{1.66}} & \best{\result{92.69}{0.90}} & \best{\result{95.29}{0.40}} & \best{\result{96.61}{0.28}} \\
\midrule
\multirow{2}{*}{Medium20} & dense-first (ours) & \result{35.71}{1.89} & \best{\result{50.32}{1.31}} & \best{\result{64.02}{0.99}} & \best{\result{72.26}{0.58}} & \best{\result{75.73}{0.74}} \\
 & sparse-first (reverse) & \best{\result{36.68}{0.69}} & \result{47.58}{1.75} & \result{62.78}{1.82} & \result{70.83}{1.44} & \result{74.62}{0.75} \\
\midrule
\multirow{2}{*}{Hard20} & dense-first (ours) & \best{\result{27.61}{3.29}} & \best{\result{35.03}{1.53}} & \best{\result{42.96}{0.51}} & \best{\result{51.14}{0.80}} & \best{\result{55.12}{0.92}} \\
 & sparse-first (reverse) & \result{26.41}{1.49} & \result{32.30}{1.84} & \result{41.45}{1.46} & \result{50.41}{0.63} & \result{53.64}{1.61} \\
\midrule
\multirow{2}{*}{Mixed30} & dense-first (ours) & \best{\result{48.55}{1.22}} & \result{54.23}{0.70} & \best{\result{62.68}{1.23}} & \result{68.43}{0.67} & \best{\result{71.69}{0.30}} \\
 & sparse-first (reverse) & \result{47.55}{1.65} & \best{\result{55.04}{0.81}} & \result{61.07}{0.64} & \best{\result{68.61}{0.32}} & \result{70.87}{0.54} \\
\midrule
\multirow{2}{*}{Full49} & dense-first (ours) & \best{\result{40.32}{0.82}} & \best{\result{47.90}{2.57}} & \best{\result{56.43}{1.75}} & \best{\result{64.57}{1.17}} & \best{\result{68.59}{0.56}} \\
 & sparse-first (reverse) & \result{39.41}{1.01} & \result{45.14}{1.79} & \result{54.80}{1.46} & \result{63.44}{0.94} & \result{67.36}{0.39} \\
\bottomrule
\end{tabular}%
}
\end{table}

\noindent\textbf{TTM ablation.} TTM trains each student with hard-label supervision and KL matching to the soft predictions of a frozen teacher. Because all reported distilled-set results use this teacher-assisted objective, we ablate its two terms to clarify how the supervision protocol affects accuracy rather than attributing the results solely to data construction. Adding KL matching to label-only supervision improves accuracy by $3.6$ to $5.2$ points, whereas adding label supervision to KL-only training changes it by only $-0.4$ to $1.0$ points. This asymmetry suggests that soft teacher targets are the main contributor within the tested TTM settings.

\subsection{Qualitative Analysis}
\label{sec:analysis}

\begin{figure}[!t]
    \centering
    \includegraphics[width=\linewidth]{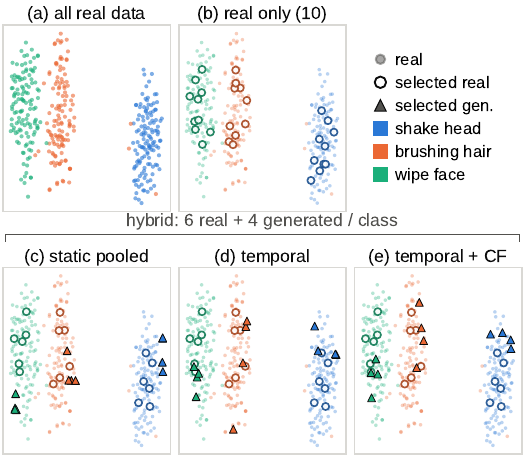}
    \caption{Shared t-SNE visualization of three action classes: (a) all real training sequences, (b) real-only selection at SPC $=10$, and (c)--(e) hybrid sets with six real and four generated sequences per class. The hybrid sets use (c) static pooled conditioning, (d) temporal conditioning, and (e) temporal conditioning with confidence filtering (CF). Small dots show real sequences, open circles mark selected real sequences, and triangles mark generated sequences. Blue, orange, and green denote shake head, brushing hair, and wipe face, respectively. All panels use the same projection of globally pooled 256-D classifier features.}
    \label{fig:qual_tsne}
\end{figure}

\figref{fig:qual_tsne} compares the full real-data distribution with compact sets for three actions. Panel (a) shows all real training sequences, and panel (b) shows ten selected real sequences per class. Panels (c)--(e) keep the same budget of ten sequences per class but combine six selected real sequences with four generated ones. The open circles and triangles show which samples come from each source.

Compared with the real-only set in panel (b), the hybrid sets use four generated samples in place of four real ones at the same per-class budget. In panels (c)--(e), the triangles often reach gaps or outer parts of the class distributions that are less represented by the real-only selections. This broader spread is visible near the lower edge of the green class, in the lower orange tail in panel (d), and toward the upper blue cluster in panel (e). The generated samples therefore appear to extend the selected range, although the effect varies across classes and conditioning variants. Some triangles also lie beyond the full real-data clouds, so a wider spread may reflect outliers rather than useful coverage. It is also worth pointing out that these observations are qualitative: t-SNE can distort distances and neighborhood structure, so the figure cannot establish coverage or class boundaries in the original feature space. It also cannot directly show that generated samples alone cause the accuracy differences.

\section{Conclusion}
We introduced \method, an integrated skeleton-based dataset distillation framework that combines motion diffusion conditioned on temporally resolved class prototypes, motion-aware training, and budget-aware hybrid selection, and evaluated it using five difficulty-stratified subsets of NTU RGB+D 60. On Full49 at SPC $=3$, \method achieves $40.32\%$ top-1 accuracy, 4.57 percentage points above class-wise $k$-means, and the same distilled sets train ST-GCN, MS-G3D, and CTR-GCN without regeneration. Our ablations also show that motion quality and recognition accuracy do not always improve together: temporal losses reduce normalized jerk with little change in accuracy, while explicit bone-length matching lowers BLV but can reduce accuracy. As the number of sequences per class increases, the accuracy gap between \method and real-only selection narrows. At larger budgets, real-only selection reaches similar accuracy with lower construction cost. One possible reason is that our current rule increases both the number and share of real samples as the budget grows. These real examples then cover more of the feature space, leaving less new coverage for generated samples to add. Future work will adapt the mix of real and generated samples to each budget and select generated examples based on the additional coverage and training benefit they provide.

\ifCLASSOPTIONcaptionsoff
  \newpage
\fi

\bibliographystyle{IEEEtran}
\bibliography{reference}
\end{document}